# Attention Fusion Reverse Distillation for Multi-Lighting Image Anomaly Detection

Yiheng Zhang, Yunkang Cao, *Student Member, IEEE*, Tianhang Zhang, Weiming Shen, *Fellow, IEEE*

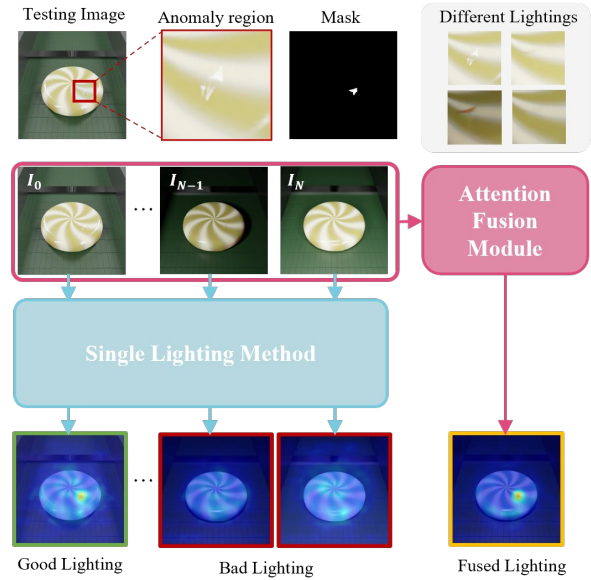

Figure 1. The scheme of the proposed method.

*Abstract*—This study targets Multi-Lighting Image Anomaly Detection (MLIAD), where multiple lighting conditions are utilized to enhance imaging quality and anomaly detection performance. While numerous image anomaly detection methods have been proposed, they lack the capacity to handle multiple inputs for a single sample, like multi-lighting images in MLIAD. Hence, this study proposes Attention Fusion Reverse Distillation (AFRD) to handle multiple inputs in MLIAD. For this purpose, AFRD utilizes a pre-trained teacher network to extract features from multiple inputs. Then these features are aggregated into fused features through an attention module. Subsequently, a corresponding student net-work is utilized to regress the attention fused features. The regression errors are denoted as anomaly scores during inference. Experiments on Eyecandies demonstrates that AFRD achieves superior MLIAD performance than other MLIAD alternatives, also highlighting the benefit of using multiple lighting conditions for anomaly detection.

## I. Introduction

Image anomaly detection aims to identify irregularities that deviate from the norm, presenting as a crucial step in ensuring product quality for industrial production [1], [2], [24]. Although significant advancements have been made in the field of image anomaly detection, the impact of various physical factors on the detection efficacy remains underexplored in the current literature.

This study particularly focuses on the impact of lighting conditions on image anomaly detection systems. Optimal lighting is essential for acquiring high-quality images that allow for the clear detection of anomalies. However, due to the complexity of object surfaces and the unpredictable nature of anomalies, relying on a single lighting condition for optimal imaging is impractical. In fact, anomalies may not be evident under certain lighting conditions but become more distinguishable under others, as illustrated in Fig.1. Therefore, this study adopts a Multi-Lighting Image Anomaly Detection (MLIAD) setting [16], where samples are imaged under multiple lighting conditions, and images captured under various lighting conditions are utilized for anomaly detection on the given samples. MLIAD proves to be a viable setting because lighting conditions can change rapidly, and capturing multi-lighting images brings slightly longer acquisition time while substantially enhancing the imaging quality of defects.

However, the introduction of multi-lighting brings the challenge of multiple inputs and requires an effective strategy for aggregating multi-lighting information for a cohesive analysis. Despite advancements in image anomaly detection techniques, most existing methods are tailored for singular lighting conditions. This narrow focus neglects the reality of varying environmental conditions objects face in real-world settings. While these methods may perform well under controlled lighting, their accuracy could diminish in the face of multi-lighting's complexity and variability, highlighting a gap in the anomaly detection landscape.

In response to this gap, this paper introduces the Attention Fusion Reverse Distillation (AFRD) method. AFRD is

*Resrach supported by the Fundamental Research Funds for the Central Universities: HUST:2021GCRC058.

Yiheng Zhang, Yunkang Cao, Tianhang Zhang and Weiming Shen are with State Key Laboratory of Digital Manufacturing Equipment and Technology, School of Mechanical Science and Engineering, Huazhong University of Science and Technology, Wuhan, China. Their emails are yiheng-zhang@hust.edu.cn, cyk_hust@hust.edu.cn, m202370548@hust.edu.cn, wshen@ieee.org (*corresponding author).

designed to fuse multi-lighting information effectively through a RD paradigm and an attention module. The architecture encompasses a pre-trained teacher encoder that extracts features across various lighting conditions. These features are then fused via the attention module, emphasizing the most critical information for anomaly detection. The attention mechanism assigns weights to features from multiple lighting conditions, ensuring the focus is on those most indicative of anomalies. A student decoder subsequently regresses the multi-lighting features. Through this intricate architecture, AFRD enhances the anomaly detection process by focusing on the most informative features across various lighting conditions.

The contributions of this paper are summarized as follows:

1) It underscores the importance of lighting in image anomaly detection and articulates the need for a multi-lighting approach, presenting the problem formulation for multi-lighting and comparing multi-lighting settings against single-lighting benchmarks to highlight its superiority in revealing otherwise obscured anomalies.

2) It presents AFRD, a novel method that addresses the complexities of multi-lighting image anomaly detection by harnessing attention fusion and reverse distillation. Through extensive experiments on Eyecandies [16] dataset, AFRD is demonstrated to achieve state-of-the-art performance, with an AUROC of 0.861, thereby setting new benchmarks for the field.

By exploring the potential of multi-lighting in enhancing image anomaly detection and proposing a viable solution to its inherent challenges, this study aims to pave the way for future research and application in this domain.

The rest of this paper is organized as follows. Section II reviews the related work. The proposed AFRD method is presented in Section III. The experiments are discussed in Section IV. Section V gives the conclusion of this study.

## II. RELATED WORK

The proposed method is related to two topics, *i.e.*, image anomaly detection and adverse imaging factors.

### A. Image Anomaly Detection

Image anomaly detection techniques are broadly categorized into two types: reconstruction-based and density-based approaches [13].

Reconstruction-based approaches focus on only reconstructing the normal images regardless of whether the input is abnormal or normal, identifying anomalies through the discrepancies observed between the original and the reconstructed images. Early strategies [9] introduce an SSIM loss to enhance the quality of reconstruction. In addition to improving the loss function, RIAD [3] approach is developed to better reconstruction by incorporating a repainting task. Despite the advancements in this approach, they still produce some blurry reconstructions, causing inaccurate detection.

On the other hand, density-based techniques concentrate on characterizing the distribution of normal features via pre-trained deep neural network. These methods achieve anomaly detection by distinguishing distances between extracted features. For example, SPADE [4] leverages pre-trained ResNet and KNN for quick, training-free pixel-level anomaly detection, enhancing accuracy by identifying mismatches with normal images. PaDiM [11] enhances this by modeling multi-level semantic correlations with Gaussian distributions for precise anomaly localization. PatchCore [7], on the other hand, focuses on maximizing test-time information from normal image patches through memory-efficient coreset subsampling.

Beyond the density-based methods mentioned, knowledge distillation methods stand out for their ability to effectively model normal feature distribution, leveraging the strengths of knowledge distillation to enhance anomaly detection. One of the initial methods, ST [5], [23], detects anomaly by aligning features between a pre-trained teacher network and a newly initialized student network. Given the uniform nature of normal images, this approach faced challenges due to the student network's potential for overfitting. The RD [6] introduces a novel reversed distillation architecture to mitigate the risk of overgeneralization by focusing on high-level representations before low-level features. Therefore, it achieves better performance and the proposed AFRD method adopts this framework for anomaly detection.

### B. Adverse Imaging Factors

In the field of anomaly detection, several factors can significantly impact imaging quality, thereby affecting the accuracy and efficiency of anomaly detection models. These factors include lighting conditions, camera angles, resolution, and environmental variables. Each of these elements can either enhance or obscure the visibility of anomalies, making their detection more challenging under non-ideal conditions.

Among these factors, lighting conditions emerge as the primary determinant of image quality, casting shadows, creating glare, or obscuring details critical for identifying anomalies. Similarly, the angle and positioning of cameras can either reveal or conceal defects, while environmental conditions like fog, dust, or smoke may reduce visibility and image clarity. The resolution of captured images further dictates the level of detail available for analysis, making high-resolution imagery a prerequisite for detecting minute anomalies. Moreover, existing research has begun to acknowledge the potential of multi-lighting data, proposing datasets and viewpoints that advocate for its utility in enhancing anomaly detection. The VISION Datasets [18] notes its lack of lighting variability and acknowledges the need for more diverse conditions in future research. The available Eyecandies [16] incorporates various lighting conditions for enhanced anomaly detection while MSC-AD [17] builds a dataset incorporating changes in illumination intensity, demonstrating the field's growing recognition of lighting's impact on anomaly detection, despite this dataset not being publicly available.

Responding to these challenges, only a few advancements have been proposed towards developing adaptive and robust anomaly detection models. HIT-MiLF [15] specifically targets non-ideal images in industrial settings, enhancing detection capabilities by simulating various conditions that affect image quality. Leroux et al. [14] integrates Multi-Branch Neural Networks in video anomaly detection under adverse lighting conditions to overcome challenges posed by lighting variations.

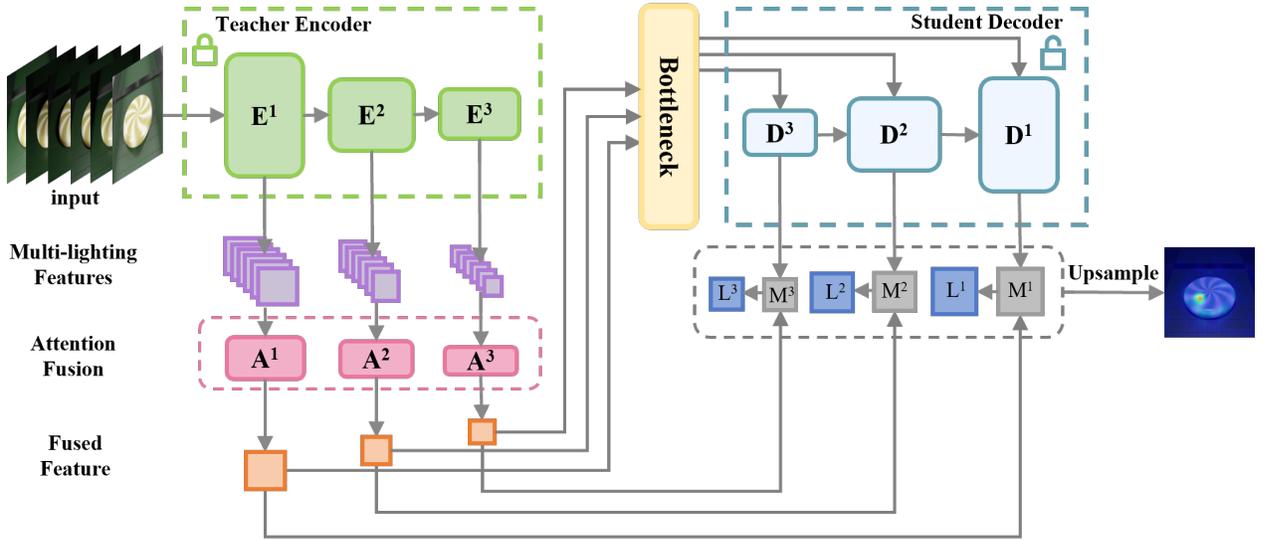

Figure 2. The overall framework of the proposed AFRD.

In summary, while numerous datasets have been proposed to tackle anomaly detection under diverse imaging conditions, particularly in MLIAD setting, there remains a notable scarcity of methods designed to utilize multiple lighting data. Therefore, this study introduces AFRD, proposing a novel method to leverage multi-lighting information effectively and pave the way for optimizing anomaly detection with adverse imaging factors.

## III. THE PROPOSED METHOD

**Problem formulation:** In the context of image anomaly detection under multi lighting conditions, each object is represented by a set of images captured under different lighting conditions. Each ensemble of lighting-condition images $I_n^j$, $j \in \{0, N\}$ constitutes an image set $S_n$ for a single object, N denotes the number of lighting conditions. Sets of anomaly-free images $\{S^t\}$ are provided for training a model to detect anomalies within unseen image sets, while sets $\{S^q\}$ contain both normal and abnormal samples are provided for testing.

The goal is to utilize the combined information from the image set to train a model that effectively identifies and localizes anomalies for an object. The focus is on interpreting the set as a whole, not separate images, to improve anomaly detection accuracy across lighting variations.

**Method Overview:** As shown in Fig.2, the AFRD method consist of four modules: a fixed pre-trained teacher encoder, a trainable bottleneck module, a trainable student decoder, and an attention fusion module for multi-lighting features fusion. The teacher encoder, student decoder and bottleneck form the reverse distillation paradigm [6] to detect anomaly. During training the student only learn the anomaly-free data from the teacher, therefore anomaly features distinguish significantly during inference procedure. Bottleneck module feeds low-level features to the student decoder for better decoding from the Teacher's knowledge while the attention module is designed to fuse features. Firstly, the pre-trained teacher encoder extracts features from different lightings, the attention module then integrates the features and pass them to the bottleneck, emphasizing relevant information for detecting anomalies. The student decoder subsequently regresses the features to calculate anomaly score.

### A. Reverse Distillation Paradigm

This study follows the reversed distillation paradigm, the input changes to a set of different lighting images $S_n$. In the paradigm, the teacher encoder E is a pre-trained encoder on ImageNet [21] used to encode representations from inputs and the parameters of it are frozen during knowledge distillation. Denote the teacher encoder as $E_T$ and a set of multi-lighting features $F_n^j$, $j \in \{0, N\}$ from multi-lighting images is extracted, N here denotes the number of lighting conditions,

$$F_n^j = E_T(S_n) \quad (1)$$

Then the obtained multi-lighting features are fused into one feature $F_f$, by the attention fusion module $A$,

$$F_f = A(F_n^j) \quad (2)$$

In the reverse distillation paradigm student decoder D is designed to restore the presentations of a teacher encoder E. However, the Teacher's activation output is not directly feed to the student. This is because the teacher's complexity might lead to unnecessary and intricate representations that overwhelm the student learning of anomaly-free features.

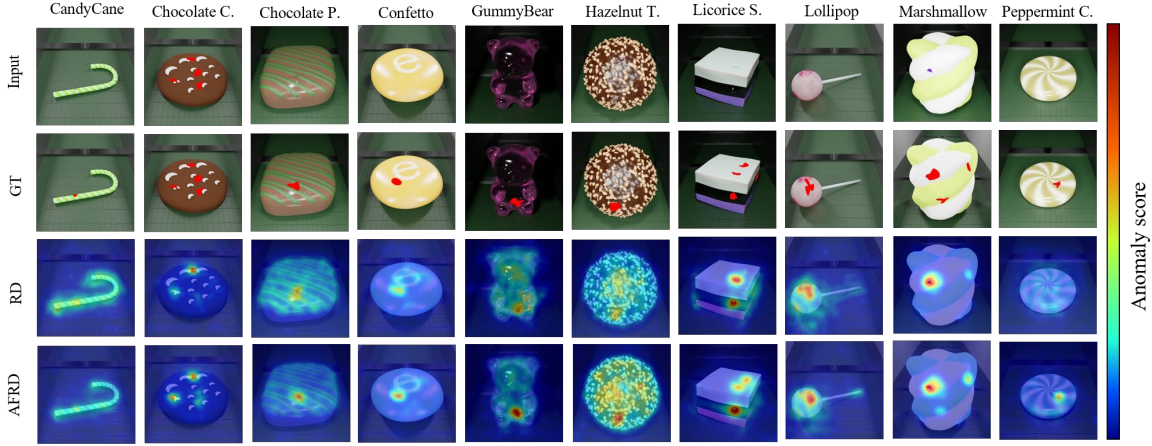

Figure 4. Qualitative comparisons between the proposed AFRD and the best single lighting method RD.

Therefore, the bottleneck is introduced to project the fused feature $F_f$ to compact embedding space $F_b$.

$$F_b = B(F_f) \qquad (3)$$

The student decoder $D_s$ structed as a reversed counterpart to teacher encoder $E_T$ then decodes the representation from the compact embedding space $F_b$ into the student feature $F_d$.

$$F_d = D_s(F_b) \qquad (4)$$

In the AFRD, the student decoder D is designed to mimic the output of the attention fused feature $F_f$ during training. And the cosine similarity loss is used for knowledge distillation.

$$\mathcal{L} = 1 - \frac{\left(F_f(h,w)\right)^T \cdot F_d(h,w)}{\parallel F_f(h,w) \parallel \parallel F_d(h,w) \parallel} \qquad (5)$$

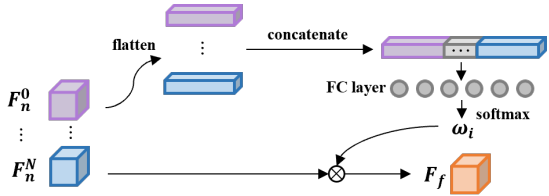

Figure 3. The attention fusion module for in the AFRD.

### B. Attention Fusion

The attention fusion module in AFRD selectively merges features across different lighting conditions, focusing on the most relevant information for effective anomaly detection, as shown in Fig.3.

The attention mechanism employed in the proposed method operates through a fully connected layer that computes attention weights from the flattened features of the input set. It inputs a set of features $F_n^j$, then uses a fully connected layer to flattens and concatenates these features.

$$\omega = F_{Softmax}\left(F_{FC}\left(F_{Concat}\left(F_{Flatten}(F_n^j)\right)\right)\right) \qquad (6)$$

These weights are then applied to the features, allowing the model to selectively emphasize information that is more indicative of normal patterns.

$$F_f = \sum_{j=0}^{f} \omega_j \cdot F_n^j, j \in \{0, N\} \qquad (7)$$

The attention fusion module effectively integrates the features from different lightings, accentuating crucial details that contribute to the accurate detection and localization of anomalies within the multi-lighting framework.

## IV. EXPERIMENTS AND DISCUSSIONS

This research conducts thorough experiments to evaluate the effectiveness of the proposed AFRD, the significance of multi-lighting fusion idea and the influence of the attention fusion module. Besides, a thorough discussion about the result is given.

### A. Experimental Setting

#### 1) Dataset

This study conducts experiments on Eyecandies dataset [16]. This dataset comprises photo-realistic images of generated candies across ten diverse classes. It designs a virtual scene to resemble an industrial conveyor belt within a light box, where the objects (candies) are placed for imaging. There are eight light sources in total. Four are positioned at the corners of the light box, and the other four are arranged around the camera at equal intervals and angles, aiming at the center where the objects are placed. For each sample object, six images are captured under different lighting conditions: One image with lightings from all four box lights. Four images with lighting from each of the camera lights individually. One image with all camera lights on simultaneously. Six distinct lighting setups are provided for each object to facilitate a comprehensive analysis of anomalies under different lighting conditions.

TABLE I. QUANTITATIVE COMPARISONS(**I-AUROC**) BETWEEN THE PROPOSED METHOD AND OTHER METHODS IN THE CANDIES DATASET. BEST RESULTS ARE HIGHLIGHTED IN **BOLD**.

| Method | G [19] | DFKDE [20] | DFM [21] | AE [22] | STFPM [5] | PaDiM [11] | RD [6] | Proposed |
|---|---|---|---|---|---|---|---|---|
| Candy Cane | 0.485 | 0.539 | 0.532 | 0.527 | 0.551 | 0.531 | 0.518 | **0.627** |
| Chocolate C. | 0.512 | 0.577 | 0.776 | 0.848 | 0.654 | 0.816 | 0.882 | **0.944** |
| Chocolate P. | 0.532 | 0.482 | 0.624 | 0.772 | 0.576 | 0.821 | **0.875** | 0.867 |
| Confetto | 0.504 | 0.548 | 0.675 | 0.734 | 0.784 | 0.856 | 0.917 | 0.968 |
| Gummy Bear | 0.558 | 0.541 | 0.681 | 0.590 | 0.737 | 0.826 | 0.835 | **0.916** |
| Hazelnut T. | 0.486 | 0.492 | 0.596 | 0.508 | **0.790** | 0.727 | 0.682 | 0.726 |
| Licorice S. | 0.467 | 0.524 | 0.685 | 0.693 | 0.778 | 0.784 | 0.771 | **0.915** |
| Lollipop | 0.511 | 0.602 | 0.618 | **0.760** | 0.620 | 0.665 | 0.578 | 0.656 |
| Marshmallow | 0.481 | 0.658 | 0.964 | 0.851 | 0.840 | **0.987** | 0.968 | **0.987** |
| Peppermint C. | 0.528 | 0.591 | 0.770 | 0.730 | 0.749 | 0.924 | 0.938 | **1.000** |
| Average | 0.507 | 0.555 | 0.692 | 0.701 | 0.708 | 0.794 | 0.796 | **0.861** |

For each category, the Eyecandies provides 6000 good images(1000 objects) for training, 1200 bad images(200 objects) and 1200 good images(200 objects) for testing.

*2) Evaluation Metric*

This study uses the image-level area under the receiver operating characteristic curve (AUROC) and pixel-level AUROC [1] to evaluate the anomaly detection performance, denoted as I-AUROC and P-AUROC.

*3) Implementation Details*

In this study, the teacher encoder is a pre-trained Wide-Resnet50 [10] on the ImageNet [8]. The student decoder utilizes a reverse Wide-Resnet50 architecture. The bottleneck employs convolutional and normalization layers to adapt and compress the feature maps from the teacher network for efficient processing and learning by the student network. This study employs AdamW [12] to optimize the two students, with a learning rate of 0.0004, a batch size of 8, and training epochs of 100.

*B. Experimental Setting*

This study implements other state-of-the-art single-lighting methods, and compare with the proposed AFRD. The experimental results showcased in Table I demonstrate the superior performance of the proposed AFRD method on the Eyecandies dataset. With an impressive average AUROC of 0.861, AFRD outperforms other state-of-the-art single-lighting methods, which are evaluated under their best single lighting conditions. In contrast, AFRD leverages a fusion of multi-lighting conditions to enhance anomaly detection capabilities, indicating its robustness and effectiveness in complex imaging scenarios where variable lighting plays a critical role.

Qualitative comparisons between the proposed AFRD and the best single lighting method RD are illustrated in Fig.4. It shows that the proposed method can reduce disturbances to better locate anomalies and detect anomalies ignored due to poor lighting conditions.

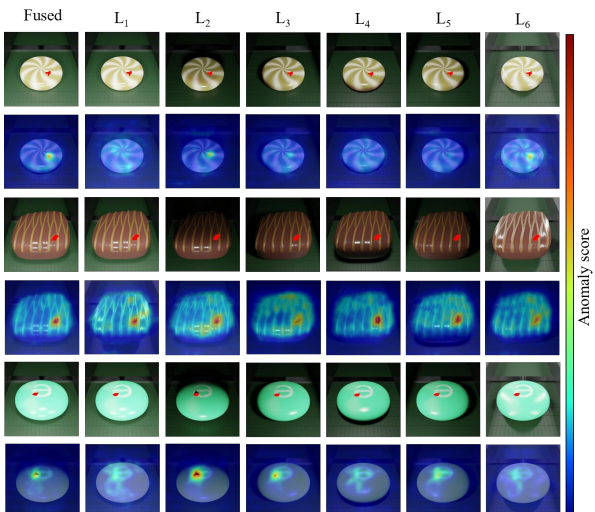

Figure 5. Qualitative comparisons using RD with different lighting conditions and with the proposed method.

TABLE II. QUANTITATIVE COMPARISONS(**I-AUROC**) BETWEEN THE EFFECTIVENESS OF DIFFERENT LIGHTINGS IN CANDIES DATASET. RD IS USED FOR COMPARISON. BEST RESULTS ARE HIGHLIGHTED IN **BOLD**.

| Category\Lighting | $L_1$ | $L_2$ | $L_3$ | $L_4$ | $L_5$ | $L_6$ |
|---|---|---|---|---|---|---|
| Candy Cane | 0.538 | 0.526 | 0.534 | **0.582** | 0.510 | 0.518 |
| Chocolate C. | 0.955 | **0.998** | 0.938 | 0.854 | 0.930 | 0.882 |
| Chocolate P. | 0.738 | 0.778 | 0.808 | 0.867 | 0.794 | **0.875** |
| Confetto | 0.920 | **0.954** | 0.923 | 0.795 | 0.925 | 0.917 |
| Gummy Bear | 0.829 | **0.893** | 0.861 | 0.798 | **0.893** | 0.835 |
| Hazelnut T. | **0.739** | 0.669 | 0.707 | 0.671 | 0.678 | 0.682 |
| Licorice S. | 0.899 | 0.885 | **0.928** | 0.874 | 0.896 | 0.771 |
| Lollipop | 0.646 | 0.643 | 0.666 | 0.622 | **0.698** | 0.578 |
| Marshmallow | 0.965 | 0.957 | 0.995 | **1.000** | 0.997 | 0.968 |
| Peppermint C. | 0.842 | **0.978** | 0.931 | 0.928 | 0.954 | 0.938 |
| Average | 0.807 | 0.828 | **0.829** | 0.799 | 0.827 | 0.796 |

*C. Significance of the Multi-lighting Fusion*

This study compares the performance of RD method under six different lightings, as shown in Table II. The experimental results under various single lighting conditions reveal that different lighting setups can significantly impact the effectiveness of anomaly detection for each object category. Moreover,

Fig.5 provides a more intuitive comparison, illustrating that the same method can yield superior results for specific object categories under certain lighting conditions compared to others. As demonstrated, some anomalies are readily detectable under one lighting condition, while the same method may struggle to identify them under a different lighting setup.

This variability indicates that certain lighting conditions can improve detection performance for specific items. Therefore, it underscores the necessity of developing methods that harness multiple lighting sources. By integrating and fusing information across different lighting scenarios, our approach aims to capitalize on the unique advantages each condition offers, leading to a more robust and comprehensive anomaly detection system.

TABLE III. QUANTITATIVE COMPARISONS BETWEEN RD, MEAN METHOD AND THE PROPOSED METHOD IN CANDIES DATASET. BEST RESULTS ARE HIGHLIGHTED IN **BOLD**.

| Metric | I-AUROC | | | P-AUROC | | |
|---|---|---|---|---|---|---|
| Method | RD | Mean | AFRD | RD | Mean | AFRD |
| Candy Cane | 0.518 | 0.578 | **0.627** | 0.961 | 0.965 | **0.971** |
| Chocolate C. | 0.882 | 0.910 | **0.944** | 0.967 | 0.979 | **0.987** |
| Chocolate P. | **0.875** | 0.842 | 0.867 | 0.960 | 0.952 | **0.976** |
| Confetto | 0.917 | 0.902 | **0.968** | 0.990 | 0.992 | **0.997** |
| Gummy Bear | 0.835 | **0.923** | 0.916 | 0.978 | 0.974 | **0.986** |
| Hazelnut T | 0.682 | 0.680 | **0.726** | 0.921 | 0.922 | **0.939** |
| Licorice S. | 0.771 | 0.902 | **0.915** | 0.955 | 0.987 | **0.992** |
| Lollipop | 0.578 | 0.632 | **0.656** | 0.978 | **0.989** | 0.985 |
| Marshmallow | 0.968 | 0.973 | **0.987** | 0.994 | 0.994 | **0.995** |
| Peppermint C. | 0.938 | 0.997 | **1.000** | 0.990 | **0.998** | 0.997 |
| Average | 0.796 | 0.834 | **0.861** | 0.969 | 0.975 | **0.982** |

### D. Influence of the Attention Fusion Module

As shown in Table III, this study conducts a comparative analysis of the RD method under single lighting condition, a simple mean fusion approach, and our proposed attention-based fusion method. The results indicate that both fusion strategies surpass the single lighting condition's best outcomes, highlighting the efficacy of multi-lighting approaches. Notably, the attention-based fusion consistently outperforms the mean fusion, validating the effectiveness and rational design of the proposed AFRD method's attention fusion module within its ablation study.

As demonstrated in the comparative analysis, the superior performance of the attention-based fusion over the mean fusion approach can be attributed to its ability to dynamically prioritize and integrate critical features from the multi-lighting inputs. This dynamic prioritization ensures that the model leverages the most discriminative features for anomaly detection, enhancing its sensitivity and specificity across varied lighting conditions. Consequently, the attention fusion module's capability to adaptively focus on pertinent features substantiates its integral role in the AFRD method, providing a targeted approach that significantly augments the model's detection capabilities.

### E. Discussions

The experimental results of this study clearly indicate that the proposed Attention Fusion Reverse Distillation method outperforms the state-of-the-art single-lighting techniques on the Eyecandies dataset, underscoring the advantage of leveraging multi-lighting conditions for anomaly detection.

In practical settings, anomaly detection often contends with varying lighting conditions, which can obscure critical details and lead to missed detections. The current research represents an initial foray into the fusion of multi-lighting information. Future methodologies could improve upon this by extracting lighting-invariant features, reducing the impact reflections, or employing more sophisticated attention mechanisms. Moreover, future work could prioritize certain lighting conditions over others, assigning variable weights on each image or even discarding less informative lighting scenarios.

Beyond just fusing information from multiple light sources, there is also potential for research into image restoration under suboptimal lighting conditions, aiming to reconstruct what the image would look like under ideal lighting. This could further enhance the robustness and accuracy of anomaly detection systems, making them more adaptable to real-world industrial settings where lighting conditions cannot always be controlled.

## V. CONCLUSION

This paper introduces AFRD method to solve the challenge of multi-lighting conditions in industrial anomaly detection, it outperforms traditional single-lighting methods by achieving an impressive average AUROC score on the Eyecandies dataset. This research underscores the pivotal role of lighting in detecting anomalies and demonstrates the potential of multi-lighting fusion to enhance detection accuracy. The study's experiments validate the efficacy of fusing multi-lighting features over single-lighting approaches and highlight the effectiveness of attention mechanisms in discerning critical features. The findings of this research pave the way for more adaptable and accurate anomaly detection in varying lighting conditions, a common scenario in production environments.

## REFERENCES

[1] P. Bergmann, K. Batzner, M. Fauser, D. Sattlegger, and C. Steger, "The MVTec Anomaly Detection Dataset: A Comprehensive Real-World Dataset for Unsupervised Anomaly Detection," *Int. J. Comput. Vis.*, vol. 129, no. 4, pp. 1038–1059, Apr. 2021.

[2] Y. Cao, X. Xu, Z. Liu, and W. Shen, "Collaborative Discrepancy Optimization for Reliable Image Anomaly Localization," *IEEE Trans. Ind. Inform.*, pp. 1–10, 2023.

[3] V. Zavrtanik, M. Kristan, and D. Skovcaj, "Reconstruction by inpainting for visual anomaly detection," *Pattern Recognit*, vol. 112, p. 107706, 2020.

[4] N. Cohen and Y. Hoshen, "Sub-image anomaly detection with deep pyramid correspondences," arXiv preprint arXiv:2005.02357, 2020.

[5] G. Wang, S. Han, E. Ding, and D. Huang, "Student-Teacher Feature Pyramid Matching for Anomaly Detection," in *British Machine Vision Conference*, Oct. 2021.

[6] H. Deng and X. Li, "Anomaly Detection via Reverse Distillation from One-Class Embedding," in *IEEE/CVF Conference on Computer Vision and Pattern Recognition*, 2022, pp. 9727–9736.

[7] K. Roth, L. Pemula, J. Zepeda, B. Schölkopf, T. Brox, and P. Gehler, "Towards total recall in industrial anomaly detection," in Proceedings of the IEEE/CVF Conference on Computer Vision and Pattern Recognition, 2022, pp. 14318–14328.

[8] O. Russakovsky *et al.*, "ImageNet Large Scale Visual Recognition Challenge," *Int. J. Comput. Vis.*, vol. 115, pp. 211–252, 2014.

[9] P. Bergmann, S. Löwe, M. Fauser, D. Sattlegger, and C. Steger, "Improving Unsupervised Defect Segmentation by Applying Structural Similarity to Autoencoders," in *VISIGRAPP*, 2018, pp. 372–380.

[10] K. He, X. Zhang, S. Ren, and J. Sun, "Deep Residual Learning for Image Recognition," in *IEEE Conference on Computer Vision and Pattern Recognition*, 2016, pp. 770–778.

[11] T. Defard, A. Setkov, A. Loesch, and R. Audigier, "PaDiM: A Patch Distribution Modeling Framework for Anomaly Detection and Localization," in *ICPR International Workshops and Challenges*, 2021, pp. 475–489.

[12] I. Loshchilov and F. Hutter, "Decoupled Weight Decay Regularization," in *International Conference on Learning Representations, ICLR 2019, May 6-9, 2019*, 2019.

[13] Y. Cao, Y. Zhang, and W. Shen, "High-Resolution Image Anomaly Detection via Spatiotemporal Consistency Incorporated Knowledge Distillation," in 2023 IEEE 19th International Conference on Automation Science and Engineering (CASE), IEEE, 2023, pp. 1–6

[14] S. Leroux, B. Li, and P. Simoens, "Multi-branch neural networks for video anomaly detection in adverse lighting and weather conditions," in Proceedings of the IEEE/CVF Winter Conference on Applications of Computer Vision, 2022, pp. 2358–2366.

[15] I. Farady, C. Kuo, H. Ng, and C. Lin, "Hierarchical Image Transformation and Multi-Level Features for Anomaly Defect Detection," Sensors, vol. 23, no. 2, pp. 988, 2023, MDPI.

[16] L. Bonfiglioli, M. Toschi, D. Silvestri, N. Fioraio, and D. De Gregorio, "The eyecandies dataset for unsupervised multimodal anomaly detection and localization," in Proceedings of the Asian Conference on Computer Vision, 2022, pp. 3586–3602.

[17] Q. Zhao, Y. Wang, B. Wang, J. Lin, S. Yan, W. Song, A. Liotta, J. Yu, S. Gao, and W. Zhang, "MSC-AD: A Multiscene Unsupervised Anomaly Detection Dataset for Small Defect Detection of Casting Surface," IEEE Transactions on Industrial Informatics, 2023, IEEE.

[18] H. Bai, S. Mou, T. Likhomanenko, R. G. Cinbis, O. Tuzel, P. Huang, J. Shan, J. Shi, and M. Cao, "VISION Datasets: A Benchmark for Vision-based Industrial Inspection," arXiv preprint arXiv:2306.07890, 2023.

[19] S. Akcay, A. Atapour-Abarghouei, and T. P. Breckon, "Ganomaly: Semi-supervised anomaly detection via adversarial training," in Asian Conference on Computer Vision, Springer, pp. 622–637.

[20] S. Akcay, D. Ameln, A. Vaidya, B. Lakshmanan, N. Ahuja, and U. Genc, "Anomalib: A deep learning library for anomaly detection," in 2022 IEEE International Conference on Image Processing (ICIP), IEEE, 2022, pp. 1706–1710.

[21] N. A. Ahuja, I. Ndiour, T. Kalyanpur, et al., "Probabilistic modeling of deep features for out-of-distribution and adversarial detection," arXiv preprint arXiv:1909.11786, 2019.

[22] G. Hinton and R. Salakhutdinov. Reducing the dimensionality of data with neural networks. Science, 313(5786):504–507, 2006.

[23] P. Bergmann, M. Fauser, D. Sattlegger, et al., "Uninformed students: Student-teacher anomaly detection with discriminative latent embeddings," in Proceedings of the IEEE/CVF Conference on Computer Vision and Pattern Recognition, 2020, pp. 4183–4192.

[24] Y. Cao, X. Xu, J. Zhang, Y. Cheng, X. Huang, G. Pang, and W. Shen, "A Survey on Visual Anomaly Detection: Challenge, Approach, and Prospect," arXiv preprint arXiv:2401.16402, 2024.